\documentclass[10pt,twocolumn,letterpaper]{article}

\usepackage{iccv}
\usepackage[utf8]{inputenc}
\usepackage[T1]{fontenc}
\usepackage{times}
\usepackage{epsfig}
\usepackage{graphicx}
\usepackage{amsmath}
\usepackage{amssymb}
\usepackage{multirow}
\usepackage{authblk}
\usepackage{amsfonts}
\usepackage{dsfont}

\usepackage[pagebackref=true,breaklinks=true,letterpaper=true,colorlinks,bookmarks=false]{hyperref}
\usepackage{cleveref}

\iccvfinalcopy 

\def\iccvPaperID{3386} 

\newcommand{\KL}{\mathrm{KL}}

\newcommand\norm[1]{\left\lVert#1\right\rVert}

\makeatletter
\newcommand{\printfnsymbol}[1]{%
  \textsuperscript{\@fnsymbol{#1}}%
}
\makeatother

\ificcvfinal\fi
\begin{document}

\title{Adversarial Autoencoders for Compact Representations of 3D Point Clouds}
\author[1,5]{Maciej Zamorski\thanks{Equal contribution}\thanks{maciej.zamorski@pwr.edu.pl}}
\author[1,5]{Maciej Zięba\printfnsymbol{1}\thanks{maciej.zieba@pwr.edu.pl}}
\author[1,5]{Piotr Klukowski}
\author[2,5]{Rafał Nowak}
\author[6]{Karol Kurach}
\author[3,5]{Wojciech Stokowiec\thanks{Currently working at DeepMind}}
\author[4,5]{Tomasz Trzciński}
 \affil[1]{Wrocław University of Science and Technology, Wrocław, Poland}
 \affil[2]{University of Wrocław, Wrocław, Poland}
 \affil[3]{Polish-Japanese Academy of Information Technology, Warsaw, Poland}
 \affil[4]{Warsaw University of Technology, Warsaw, Poland}
 \affil[5]{Tooploox, Wrocław, Poland}
 \affil[6]{Google Brain, Zurich, Switzerland}
\maketitle
\begin{abstract}
Deep generative architectures provide a way to model not only images but also complex, 3-dimensional objects, such as point clouds.
In this work, we present a novel method to obtain meaningful representations of 3D shapes that can be used for challenging tasks including 3D points generation, reconstruction, compression, and clustering.
Contrary to existing methods for 3D point cloud generation that train separate decoupled models for representation learning and generation, our approach is the first end-to-end solution that allows to simultaneously learn a latent space of representation and generate 3D shape out of it.
Moreover, our model is capable of learning meaningful compact binary descriptors with adversarial training conducted on a latent space.
To achieve this goal, we extend a deep Adversarial Autoencoder model (AAE) to accept 3D input and create 3D output.
Thanks to our end-to-end training regime, the resulting method called 3D Adversarial Autoencoder (3dAAE) obtains either binary or continuous latent space that covers a much wider portion of training data distribution.
Finally, our quantitative evaluation shows that 3dAAE provides state-of-the-art results for 3D points clustering and 3D object retrieval.
\end{abstract}

\begin{figure}[!th]
\centering
\includegraphics[width=0.9\linewidth]{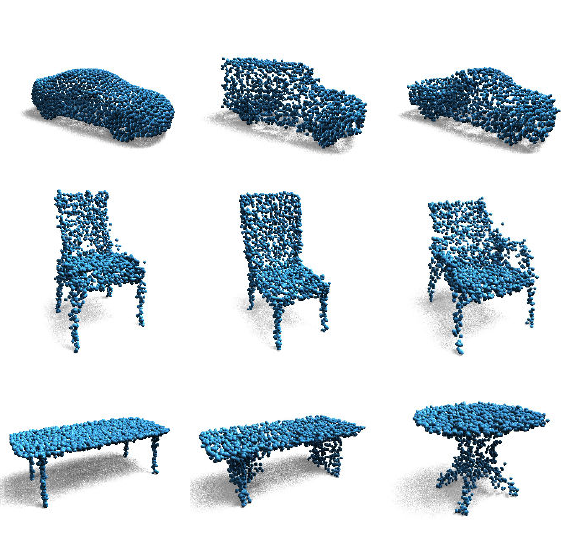}
\caption{Synthetic point cloud samples generated by our AAE models trained with Earth Mover Distance as a~reconstruction error.}
\label{fig:generated}
\end{figure}

\section{Introduction}
As more and more sensors offer capturing depth along with other visual cues, three-dimensional data points start to play a pivotal role in many real-life applications, including simultaneous localization and mapping or 3D object detection. A proliferation of devices such as RGB-D cameras and LIDARs leads to an increased amount of 3D data that is being captured and analysed, \eg~by autonomous cars and robots. Since storing and processing 3D data points in their raw form quickly becomes a bottleneck of a processing system, designing compact and efficient representations is of the utmost importance for researchers in the field \cite{gumhold2001feature,pauly2003multi,daniels2007robust,qi2017pointnet,zhou2018voxelnet,yu2018pu}.

Despite the growing popularity of point cloud representations, the labeled data in this domain is still very limited, bringing the necessity of extracting features using unsupervised \cite{shoef2019pointwise} or semi-supervised \cite{suger2015traversability} methods.
As learned features are later used for downstream tasks, such as reconstruction, generation, retrieval and clustering, the latent space is desirable to match a specified prior probability distribution. While such representations have been extensively investigated for 2D images \cite{bengio2013representation,kingma2013auto,radford2015unsupervised,tomczak2018vae,xia2014supervised,zieba2018bingan,zieba2017training}, there has been much less focus on applying these techniques to 3D shapes and structures \cite{simon2014separable}. In particular, there is no other work that obtains binary representation from point cloud data for generation, retrieval and interpolation tasks.

To address this problem, we introduce the novel single-stage end-to-end 3D point cloud probabilistic model called \emph{3D Adversarial Autoencoder} (\textbf{3dAAE}). This generative model is capable of representing 3D point clouds with compact continuous or binary embeddings. Additionally, those representations follow an arbitrary prior distribution, such as Gaussian Mixture Model (continuous embeddings), Bernoulli or Beta (binary representations).

In the proposed approach we utilize adversarial autoencoder, wherein the PointNet model \cite{qi2017pointnet} is used as an encoder. As a reconstruction loss, we use Earth-Mover distance \cite{Rubner2000,fan2017point}, which enable end-to-end training of the model with permutation invariance for input data. Finally, to guarantee the desired distribution on the latent space, we utilize adversarial training with Wasserstein criterion \cite{gulrajani2017improved}.  


In contrast to existing approaches, our model captures the point cloud data in a way, which allows for simultaneous data generation, feature extraction, clustering, and object interpolation. Interestingly, these goals are attainable by using only unsupervised training with reconstruction loss and prior distribution regularization.


To summarize, in this work we present the following contributions: $(1)$ We present comprehensive studies of variational autoencoders (VAE) and adversarial autoencoders (AAE) in context of 3D point cloud generation.
$(2)$ Although AAE outperforms VAE in our studies, we have shown that by selecting proper cost function for EMD distance we achieve proper ELBO for variational calculus.
$(3)$ We show that AAE can solve challenging tasks related to 3D points generation, retrieval, and clustering. In particular, our AAE model achieves state-of-the-art results in 3D point generation. 
$(4)$ We show our AAE model can represent 3D point clouds in compact binary space (up to 100 bits) and achieves competitive results to models that are using wide and continuous representations.

\section{Related Work}
\label{sec:related}
\textbf{Point clouds}
The complex task of creating rich 3D point clouds feature representations is an active field of investigation. In \cite{wu20153d} authors propose the voxelized representation of an input point cloud. Other approaches are using such techniques as multi-view 2D images \cite{su2015multi} or calculating occupancy grid \cite{ji20133d,maturana2015voxnet}.
The PointNet architecture \cite{qi2017pointnet} allows handling unordered sets of real-valued points, by introducing permutation-invariant feature aggregating function. This approach has constituted a basis for numerous extensions, involving training hierarchical features \cite{qi2017pointnet++}, point cloud retrieval \cite{angelina2018pointnetvlad} or point cloud generation \cite{achlioptas2018learning}. 

\textbf{Representation learning}
The studies on 3D-GAN model \cite{3dgan} have resulted in an extension of an original GAN  \cite{goodfellow2014generative}, which enabled generating realistic 3D shapes sampled from latent variable space. Another work in the field \cite{Schonberger17} uses a version of Variational Autoencoder (VAE) \cite{kingma2013auto,rezende2014stochastic} to build a semantic representation of a 3D scene that is used for relocalization in SLAM. 
However, models that incorporate VAE into the framework usually regularize the feature space with a normal distribution. 
It has been shown, that there are better prior distributions to optimize for \cite{tomczak2018vae}. 
Moreover, due to the necessity for defining a tractable and differentiable KL-divergence, VAE allows only for a narrow set of prior distributions.
As a way to apply an arbitrary prior regularization to the autoencoder, the Adversarial AutoEncoder framework has been proposed \cite{makhzani2015adversarial}. It regularizes latent distribution via discriminator output, rather than KL-divergence.
This approach has been used to 2D image data but has not been validated in the context of 3D point cloud data.

The only reference method, known to the authors, for calculating binary 3D feature descriptors is B-SHOT \cite{prakhya2015b}. The approach was designed for binarization of initial SHOT descriptor \cite{salti2014shot}, which was designed for keypoint matching task.

To sum up, numerous existing solutions are able to tackle generating or representation of 3D point clouds. However, we are lacking end-to-end solutions, which can capture both of them and allow for learning compact binary descriptors or clustering.

\section{Methods}
\label{sec:methods}

\subsection{Variational Autoencoders}

Variational Autoencoders (\textbf{VAE}) are the generative models that are capable of learning approximated data distribution by applying variational inference \cite{kingma2013auto,rezende2014stochastic}. We consider the latent stochastic space $\mathbf{z}$ and optimize the upper-bound on the negative log-likelihood of $\mathbf x$:

\begin{gather}
\begin{aligned}
\mathbb{E}_{\mathbf{x} \sim p_d(\mathbf{x})}[-\log{p(\mathbf{x})}] < & 
\mathbb{E}_{\mathbf{x}}[\mathbb{E}_{z \sim q(\mathbf{z}|\mathbf{x})}[-\log{p(\mathbf{x}|\mathbf{z})}]] \nonumber \\
& + \mathbb{E}_{\mathbf{x}}\left[\KL\left(q(\mathbf{z}|\mathbf{x}) \| p(\mathbf{z})\right)\right]
\end{aligned} \\
= \textnormal{reconstruction} \; +  \;  \textnormal{regularization},
\end{gather}
where $\KL(\cdot\|\cdot)$ is the Kullback–Leibler divergence~\cite{kullback1951information}, $p_d(\mathbf{x})$ is the empirical distribution, $q(\mathbf{z}|\mathbf{x})$ is the variational posterior (the encoder $E$), $p(\mathbf{x}|\mathbf{z})$ is the generative model (the generator $G$) and $p(\mathbf{z})$ is the prior. In practical applications $p(\mathbf{x}|\mathbf{z})$ and $q(\mathbf{z}|\mathbf{x})$ are parametrized with neural networks and sampling from $q(\mathbf{z}|\mathbf{x})$ is performed by so called reparametrization trick. The total loss used to train VAE can be represented by two factors: reconstruction term, that is $\ell_2$ norm taken from the difference between sampled and reconstructed object if $p(\mathbf{x}|\mathbf{z})$ is assumed to be normal distribution and regularization term that forces $\mathbf{z}$ generated from $q(\mathbf{z}|\mathbf{x})$ network to be from a prior distribution $p(\mathbf{z})$.  

\begin{figure}[ht]
\begin{center}
    \includegraphics[width=0.99\linewidth]{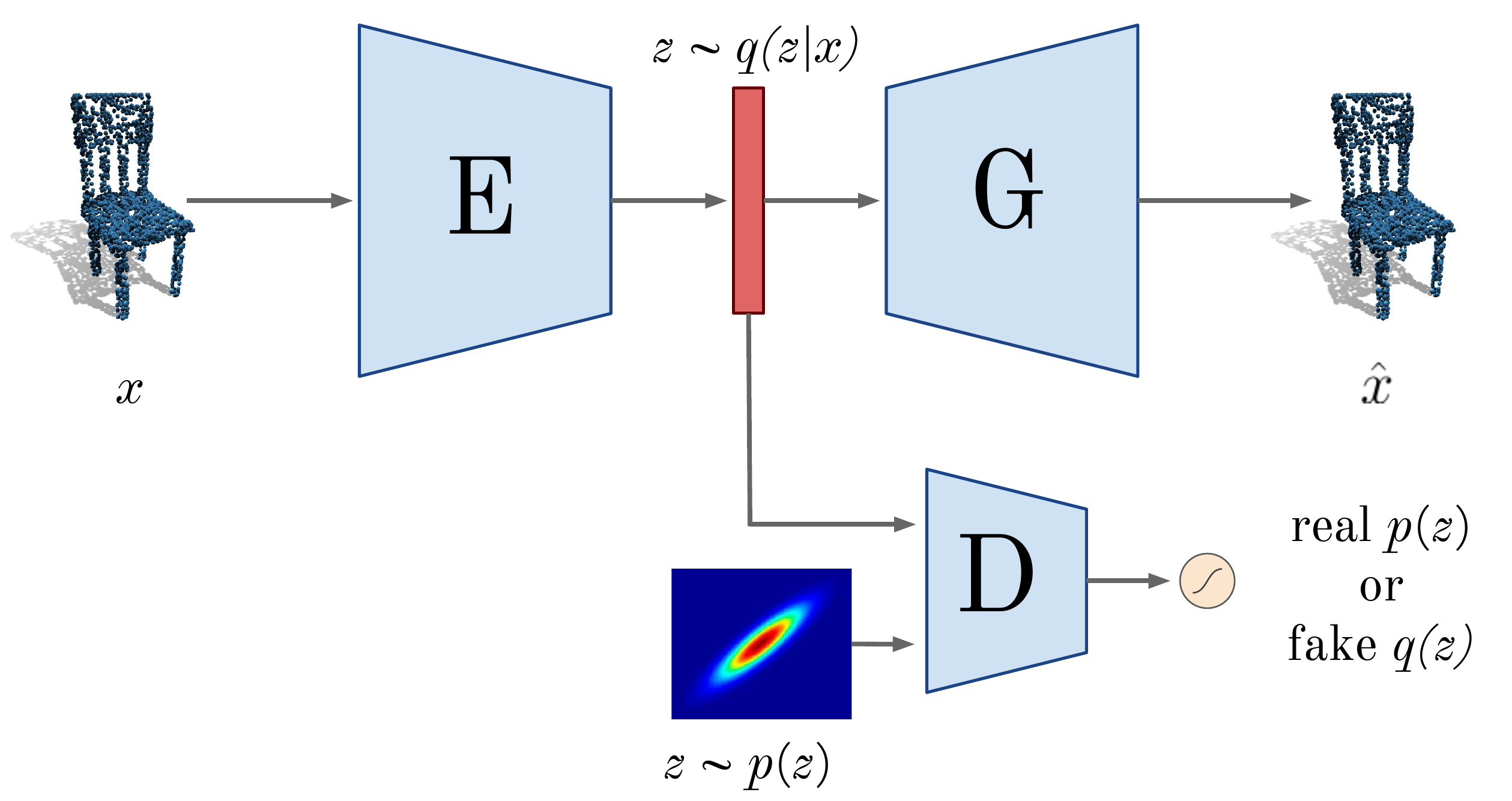}
\end{center}
\caption{3dAAE model architecture that extends AE with an additional decoder D. The role of the decoder is to distinguish between true samples generated from $p(\mathbf{z})$ and fakes delivered by the encoder E. Encoder is trying to generate artificial samples, that are difficult to be distinguished by the discriminator.}
\label{fig:avae}
\end{figure}

\subsection{Adversarial Autoencoders}
\label{sec:adversarial}

The main limitation of VAE models is that regularization term requires particular prior distribution to make $\KL$ divergence tractable. In order to deal with that limitation authors of \cite{makhzani2015adversarial} introduced Adversarial Autoencoders (\textbf{AAE}) that utilize adversarial training to force a particular distribution on $\mathbf{z}$ space. The model assumes that an additional neural network - discriminator $D$, which is responsible for distinguishing between fake and true samples, where the true samples are sampled from assumed prior distribution $p(\mathbf{z})$ and fake samples are generated via encoding network $q(\mathbf{z}|\mathbf{x})$. The adversarial part of training can be expressed in the following way: 
\begin{multline}
\min_{E} \max_{D} V(E,D) =\mathbb{E}_{\mathbf{z} \sim p(\mathbf{z})}{[\log{D(\mathbf{z})}]} \\ 
+ \mathbb{E}_{\mathbf{x} \sim p_d(\mathbf{x})}{[\log{(1-D(E(\mathbf{x})))}]}.
\end{multline}

The training procedure is characteristic for GAN models and is performed by alternating updates of parameters of encoder $E$ and discriminator $D$. 
The parameters of discriminator $D$ are updated by minimizing the $L_D=-V(E,D)$ and the parameters of encoder $E$ and generator $G$ are optimized by minimizing the reconstruction error together with $V(E,D)$: $L_{EG} = reconstruction + V(E,D)$. 
In practical applications, the stated criterion can be substituted with so-called Wasserstein criterion \cite{gulrajani2017improved}. 
Other approaches that are found in the literature that apply Wasserstein criterion are Wasserstein Autoencoders \cite{tolstikhin2017wasserstein} (of which Adversarial Autoencoders are a special case) and Wasserstein VAE \cite{ambrogioni2018wasserstein}.

Learning prior on $\mathbf{z}$ latent space using adversarial training has several advantages over standard VAE approaches \cite{makhzani2015adversarial}. First of all, the data examples coded with the encoder exhibits sharp transitions indicating that the coding space is filled which is beneficial in terms of interpolating on the latent space. Secondly, there is no limitation for the distribution that is adjusted to $\mathbf{z}$ space.




\section{Our Approach}
\label{sec:aae}

VAE and AAE are widely applied for analysis of images, but there is limited work on their applications to the generation and representation of the point clouds. In our study, we have adjusted these models to address multiple challenging tasks, arising in point cloud analysis, such as generation, representation, learning binary descriptors and clustering. 

The point cloud can be represented as a set of points in 3D Euclidean space, denoted by $\mathcal{S}=\{ \mathbf{x}_n \}_{n=1}^N$, where $\mathbf{x}_n\in \mathbb{R}^3$ is a single point in 3D space. For this particular data representation, the crucial step is to define proper reconstruction loss that can be further used in the autoencoding framework. In the literature, two common distance measures are successively applied for reconstruction purposes: \textit{Earth Mover's (Wasserstein) Distance} \cite{Rubner2000} and \textit{Chamfer pseudo-distance}. 

\textit{Earth Mover's Distance} (\textbf{EMD}): introduced in \cite{Rubner2000} is a metric between two distributions based on the minimal cost that must be paid to transform one distribution into the other. For two equally sized subsets $ \mathcal{S}_1 \subseteq \mathbb{R}^3, \; \mathcal{S}_2 \subseteq \mathbb{R}^3$, their EMD is defined as:
\begin{equation}
    EMD(\mathcal{S}_1, \mathcal{S}_2) = \min_{\phi : \mathcal{S}_1 \rightarrow \mathcal{S}_2} \sum_{\mathbf{x} \in \mathcal{S}_1} c(\mathbf{x},\phi(\mathbf{x})),
    \label{eq:EMD}
\end{equation}
where $\phi$ is a bijection and $c(\mathbf{x},\phi(\mathbf{x}))$ is cost function and can be defined as $c(\mathbf{x},\phi(\mathbf{x}))=\frac{1}{2}\norm{\mathbf{x} - \phi(\mathbf{x})}_2^2$. 

\textit{Chamfer pseudo-distance} (\textbf{CD}): measures the squared distance between each point in one set to its nearest neighbor in the other set:
\begin{equation}
    CD(\mathcal{S}_1, \mathcal{S}_2) = \sum_{\mathbf{x} \in \mathcal{S}_1} \min_{\mathbf{y} \in \mathcal{S}_2} \norm{\mathbf{x} - \mathbf{y}}_2^2 + \sum_{\mathbf{y} \in  \mathcal{S}_2} \min_{\mathbf{x} \in \mathcal{S}_1} \norm{\mathbf{x} - \mathbf{y}}_2^2.
\end{equation} In contrast to EMD which is only differentiable almost everywhere, CD is fully differentiable. Additionally, CD is computationally less requiring. On the other hand, the EMD distance guarantees one-to-one point mapping and provides better reconstruction results. 

The superior generative power of GAN models \cite{ledig2017photo} over classical autoencoders in terms of generating artificial images is mainly caused by the difficulties in defining good distance measure in data (pixel) space that is essential for the encoders. Thanks to the distance measures, like Chamfer or Earth-Mover, this limitation is no longer observed for 3D point clouds. Therefore, we introduce VAE and AAE models for 3D point clouds and call them \textbf{3dVAE} and \textbf{3dAAE}, respectively.

Following the framework for VAE introduced in Section~\ref{sec:methods} we parametrize $q(\mathbf{z}|\mathbf{x})$ with the encoding network $E(\cdot)$ and expect normal distribution with diagonal covariance matrix: $q(\mathbf{z}|\mathbf{x})=\mathcal{N}(E_{\mu}(\mathbf{x}),E_{\sigma}(\mathbf{x}))$, where $\mathbf{x}$ stores the points from $\mathcal{S}$ assuming some random ordering. Because we operate on set of points as $E(\cdot)$ we utilize the PointNet model that is invariant on the permutations, therefore we receive the same distribution for all possible orderings of points from $\mathcal{S}$. On the other hand, we make use of $G$ model to parametrize separate distributions for each of the cloud points, $\mathbf{x_1},\dots,\mathbf{x}_N$, in the following manner. We model each of point distributions $p(\mathbf{x}_j|\mathbf{z})$ with the normal distribution with the mean values returned by network $G(\cdot)$, $p(\mathbf{x}_j|\mathbf{z})=\mathcal{N}(G_i(\mathbf{z}),\mathds{1})$, where $p(\mathbf{x}_1|\mathbf{z})\dots p(\mathbf{x}_N|\mathbf{z})$ are conditionally independent, $\phi(\mathbf{x}_j)=G_i(\mathbf{z})$ and $\phi(\cdot)$ is a bijection. Because we are not able to propose permutation invariant mapping $G(\cdot)$ (as it is simple MLP model) we utilize additional function $\phi(\cdot)$ that provides one-to-one mapping for the points stored in $\mathcal{S}$. For this particular assumption the problem of training the VAE model for the point clouds (\textbf{3dVAE}) can be solved by minimizing the following criterion: 
\begin{equation}
    Q = \KL(q(\mathbf{z}|\mathbf{x})\|p(\mathbf{z})) + \min_{\phi}\sum_{i=1}^N{\frac{||\mathbf{x}_i - \phi{(\mathbf{x}_i)}||_2^2}{2}}.
\end{equation}
The training criterion for the model is composed of reconstruction error defined with Earth-Mover distance between original and reconstructed samples, and $\KL$ distance between samples generated by encoder and the samples generated from prior distribution. Samples from the encoder are obtained by application of reparametrization trick on the last layer. To balance the gap caused by the orders of magnitude between the components of the loss we scale the Earth-Mover component by multiplying it by the scaling parameter $\lambda$. To be consistent with reference methods in the experimental part we also use slightly modified unsquared cost function: $c(\mathbf{x},\phi(\mathbf{x}))=\norm{\mathbf{x} - \phi(\mathbf{x})}_2$.


\begin{figure}[h]
\centering
\includegraphics[width=0.9\linewidth]{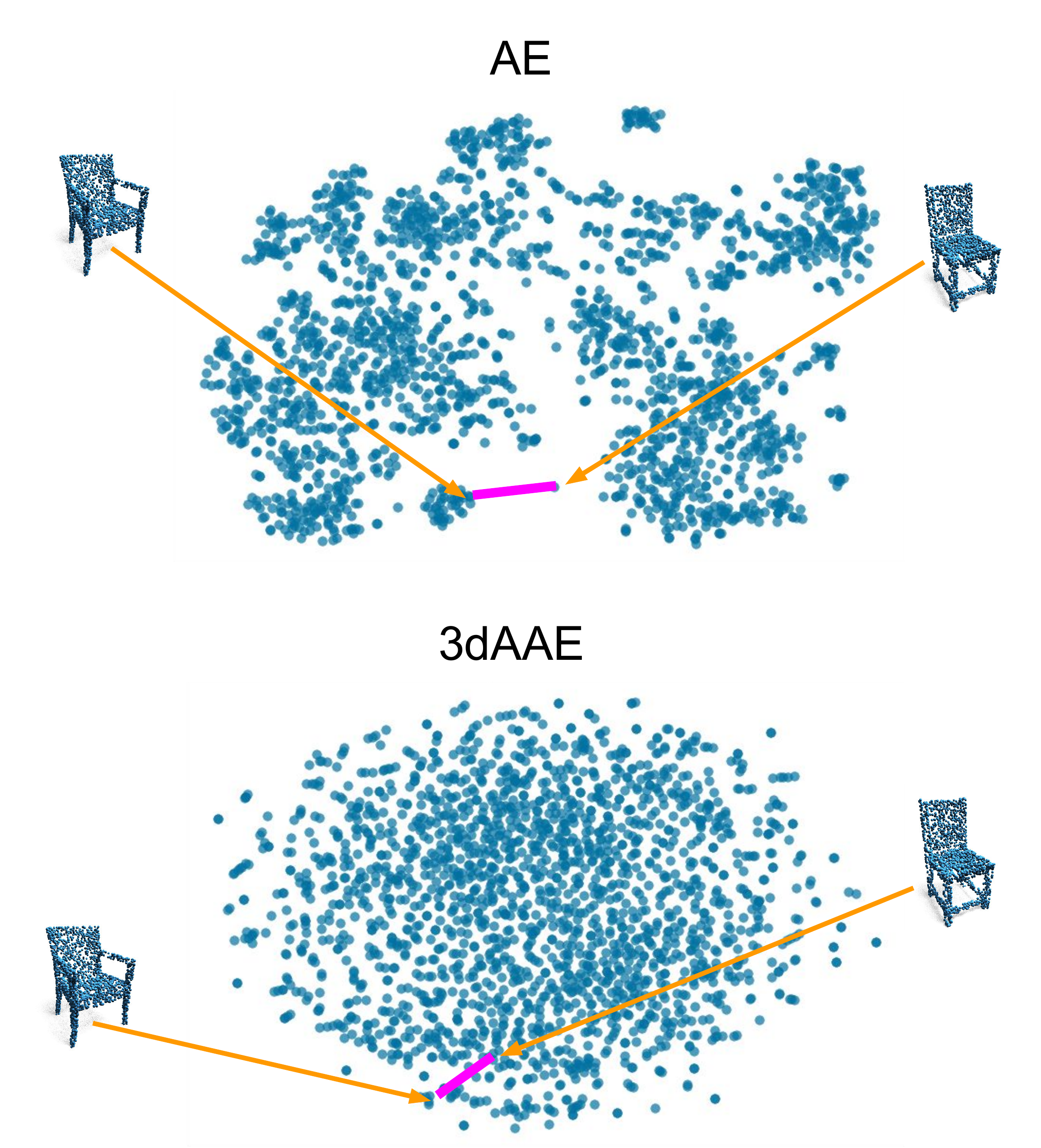}
\caption{t-SNE plot for the latent space obtained from AE and 3dAAE models (chair category). One can notice the interpolation gap between two chairs for AE. For encodings obtained from 3dAAE model this phenomenon is not observed and the latent variable space is much more dense which allows for smooth transition within the space. }
\label{fig:tsne_interpolation}
\vspace{3mm}
\includegraphics[width=0.99\linewidth]{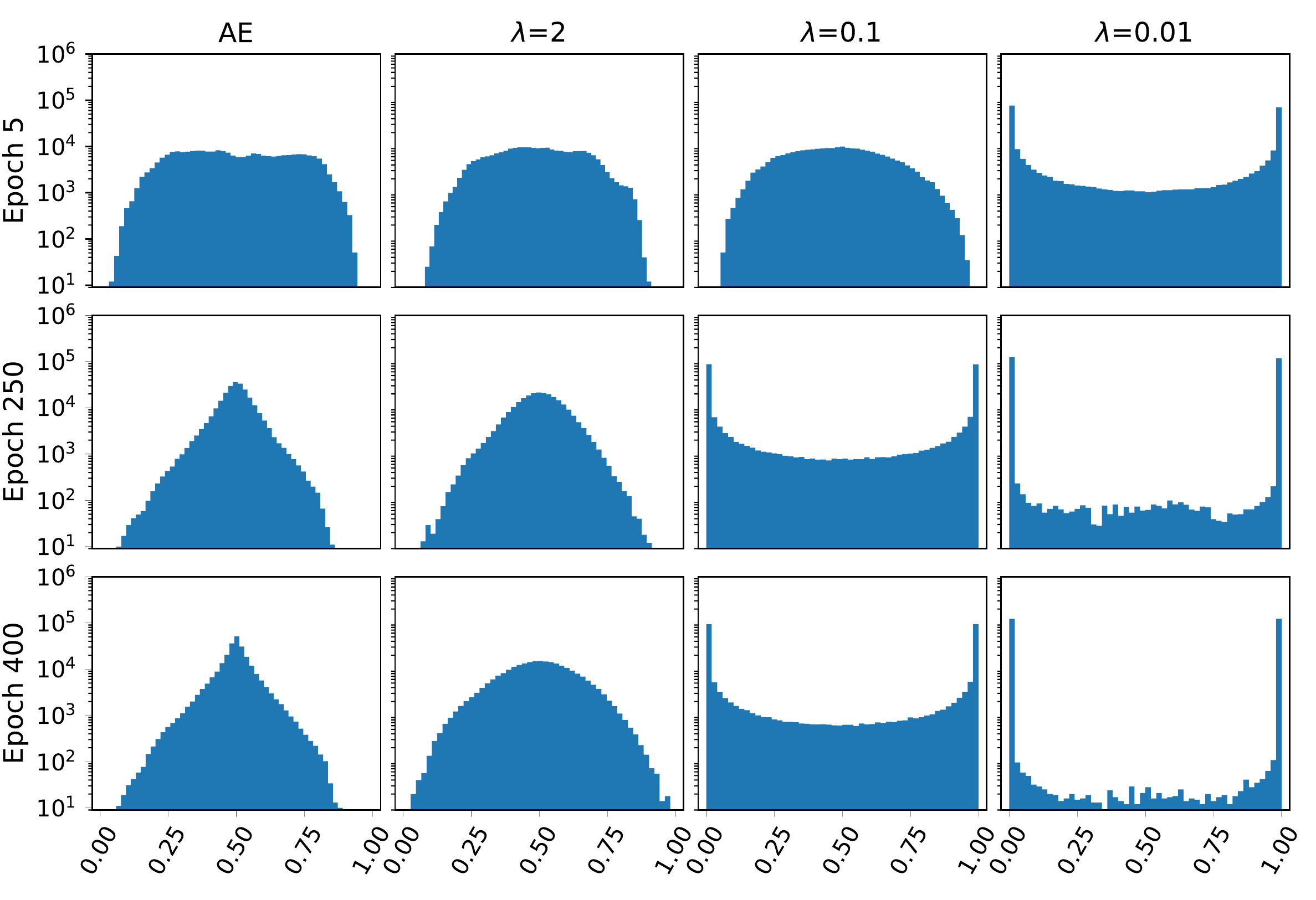}
\caption{Impact of the adversarial training with $Beta(\alpha=0.01, \beta=0.01)$ on the distribution of point cloud embeddings. To balance reconstruction and  adversarial losses, we have introduced $\lambda$ hyperparameter.}
\label{fig:distributions}
\end{figure}

\begin{figure*}[ht]
\centering
\includegraphics[width=0.9\linewidth,trim={0 0 0 2cm},clip]{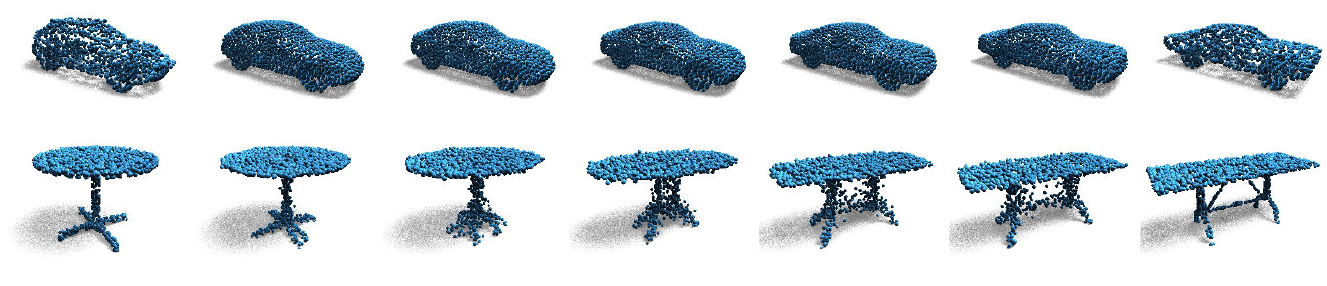}
\caption{Interpolations between the test set objects obtained by our single-class AAE models. Leftmost and rightmost samples present ground truth objects. Images in between are the result of generating images from linear interpolation between our latent space encodings of the side images.}
\label{fig:interpolations}
\end{figure*}

Due to the limitations of VAE listed in the previous section, namely narrow spectrum of possible priors and worse distribution adjustment,  we propose the adversarial approach adjusted to be applied to 3D point clouds (\textbf{3dAAE}). The scheme of adversarial autoencoder for 3D point clouds is presented in Figure~\ref{fig:avae}. The model is composed of an encoder $E$ that is represented by the PointNet \cite{qi2017pointnet}-like architecture, which transforms 3D points into latent space $\mathbf{z}$. The latent coding $\mathbf{z}$ is further used by a generator $G$ to reconstruct or generate 3D point clouds. To train the assumed prior distribution $p(\mathbf{z})$ we utilize a discriminator $D$ that is involved in the process of distinguishing between the true samples generated from the prior distribution $p(\mathbf{z})$ and the fake samples obtained from encoder $E$. If $\mathbf{z}$ is either a normal distribution or a mixture of Gaussians we apply the reparametrization trick to the encoder $E$ to obtain the samples of $\mathbf{z}$. 

The model is trained in an adversarial training scheme and described in details in Section~\ref{sec:adversarial}. As a reconstruction loss, we take Earth-Mover loss defined by eq. (\ref{eq:EMD}). For the GAN part of the training process, we utilize the Wasserstein criterion. 

In Figure~\ref{fig:tsne_interpolation} we present the 2D visualization of the coding space for AE and 3dAAE methods. For 3dAAE model, we can observe, that the encodings are clustered consistently to the prior distribution. For the AE model, we can find the gaps in some areas that may lead to poor interpolation results.  

Subsequently, we have used the proposed 3dAAE model to learn compact binary descriptors (100 bits) of 3D point clouds. The adversarial training with $Beta(0.01, 0.01)$ has allowed to alter the distribution of AE embeddings so that it accumulates its probability mass around $0$ or $1$ (Figure \ref{fig:distributions}). The binary embeddings presented in the experimental section, are obtained by thresholding $\mathbf{z}$ at 0.5.

Contrary to the stacked models presented in \cite{achlioptas2018learning}, our model is trained in an end-to-end framework, and the latent coding space is used for both representation and sampling purposes. Thanks to the application of adversarial training we obtain data codes that are consistent with the assumed prior distribution $p(\mathbf{z})$. 

\section{Evaluation}
\label{sec:evaluation}

In this section, we describe experimental results of the proposed generative models in various tasks including 3D points reconstruction, generation, a binary representation, and clustering. 

\subsection{Metrics}
Following the methodology for evaluating generative fidelity and samples diversification provided in \cite{achlioptas2018learning} we utilize the following criteria for evaluation: Jensen-Shannon Divergence, Coverage, and Minimum Matching Distance. 

\textit{Jensen-Shannon Divergence} (\textbf{JSD}): a measure of distance between two empirical distributions $P$ and $Q$, defined as:
\begin{equation}
    JSD(P || Q) = \frac{\KL(P || M) + \KL(Q || M)}{2},
\end{equation}
where $M = \frac{P + Q}{2}$.

\textit{Coverage} (\textbf{COV}): a measure of generative capabilities in terms of richness of generated samples from the model. For two point cloud sets $\mathcal{S}_1, \mathcal{S}_2\subset \mathbb{R}^3$ coverage is defined as a~fraction of points in $\mathcal{S}_2$ that are in the given metric the nearest neighbor to some points in $\mathcal{S}_1$.

\textit{Minimum Matching Distance} (\textbf{MMD}): Since COV only takes the closest point clouds into account and does not depend on the distance between the matchings additional metric was introduced. For point cloud sets $\mathcal{S}_1$, $\mathcal{S}_2$ MMD is a~measure of similarity between point clouds in $\mathcal{S}_1$ to those in $\mathcal{S}_2$. 

Both COV and MMD can be calculated using Chamfer (\textbf{COV-CD}, \textbf{MMD-CD}) and Earth-Mover (\textbf{COV-EMD}, \textbf{MMD-EMD}) distances, respectively. For completeness we report all possible combinations. 

\subsection{Network architecture}

In our experiments we use the following network architectures:
\textit{Encoder} (\textbf{E}) is a PointNet-like network composed of five \emph{conv1d} layers, one fully-connected layer and two separate fully-connected layers for reparametrization trick. ReLU activations are used for all except the last layer used for reparametrization.
\textit{Generator} (\textbf{G}) is a fully-connected network with 5 layers and ReLU activations except the last layer.  
\textit{Discriminator} (\textbf{D}) is a  fully-connected network with 5 layers and ReLU activations except the last layer. 

The above architecture is trained using Adam \cite{kingma2014adam} with parameters $\alpha = 10^{-4}$, $\beta_1=0.5$ and $\beta_2 = 0.999$.

\subsection{Experimental setup}

To perform experiments reported in this work, we have used either \emph{ShapeNet}\cite{chang2015shapenet}  or \emph{ModelNet40} \cite{wu20153d} datasets transformed to the $3\times2048$ point cloud representation following the methodology provided in \cite{achlioptas2018learning}.  Unless otherwise stated, we train models with point clouds from a~single object class and work with train/validation/test sets of an $85\%-5\%-10\%$ split. When reporting JSD measurements we use a $28^3$ regular voxel grid to compute the statistics.

\begin{figure*}[th!]
\centering
\includegraphics[width=0.9\linewidth]{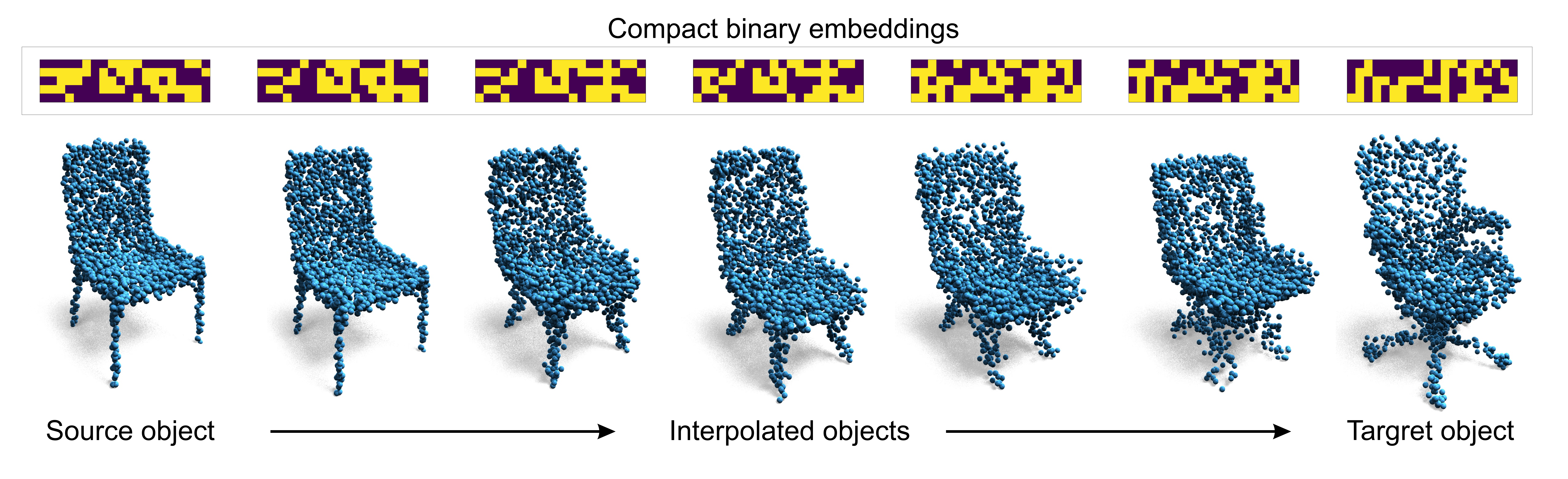}
\caption{Compact binary representations (100 bits) of 3D point clouds. For each of the 3D shapes, we provide corresponding binary codes.}
\label{fig:binary_algebra}
\end{figure*}

\subsection{Models used for evaluation}
Following the previous studies \cite{achlioptas2018learning}, we have included the following models in our work:

\textit{Autoencoder} (\textbf{AE}). The simple architecture of autoencoder that converts the input point cloud to the bottleneck representation $\mathbf{z}$ with an encoder. The model is used for representing 3D points in latent space without any additional mechanisms that can be used to sample artificial 3D objects. Two approaches to train AE are considered: 1) with Chamfer (\textbf{AE-CD}) or 2) Earth-Mover (\textbf{AE-EMD}) distance as a way to calculate reconstruction error. 

\textit{Raw point cloud GAN} (\textbf{r-GAN)}. The basic architecture of GAN that learns to generate point clouds directly from the sampled latent vector.  

\textit{Latent-space (Wasserstein) GAN} (\textbf{l-(W)GAN)}. An extended version of GAN trained in stacked mode on latent space $\mathbf{z}$ with and without an application of Wasserstein criterion \cite{arjovsky2017wasserstein}. 

\textit{Gaussian mixture model} (\textbf{GMM)}. Gaussian Mixture of Models fitted on the latent space of an encoder in the stacked mode. 

Finally, we also evaluate the models proposed in this work as well as their variations:

\textit{3D Variational Autoencoder} (\textbf{3dVAE)}. Autoencoder with EMD reconstruction error and $\KL(\cdot\|\cdot)$ regularizer.   

\textit{3D Adversarial Autoencoder} (\textbf{3dAAE}). Adversarial autoencoder introduced in Section~\ref{sec:aae} that makes use of a prior $p(\mathbf{z})$ to learn the distribution directly on a latent space $\mathbf{z}$. Various types of priors are considered in our experiments: normal distribution (\textbf{3dAAE}), mixture of Gaussians (\textbf{3dAAE-G}), beta (\textbf{3dAAE-B}).  

\textit{3D Categorical Adversarial Autoencoder} (\textbf{3dAAE-C}). AAE model with an additional category output returned by an encoder for clustering purposes (see Section~\ref{sec:CAAE}). 

\subsection{Reconstruction capabilities}

In this experiment, we evaluate the reconstruction capabilities of the proposed autoencoders using unseen test examples. We confront the reconstruction results obtained by the AE model with our approaches to examine the influence of a prior regularization on a reconstruction quality. In Table~\ref{tab:rec} we report the MMD-CD and MMD-EMD between reconstructed point clouds and their corresponding ground-truth in the test dataset of the chair object class. It can be observed, that the 3dVAE model does not suffer from overregularization problem \cite{achlioptas2018learning}. Both reconstruction measures are on a comparable level or even slightly lower for 3dAAE-G. 

\begin{table}
\begin{center}
\begin{tabular}{l|c|c}
\textbf{Method} &\textbf{MMD-CD} & \textbf{MMD-EMD}\\
\hline
\hline
AE \cite{achlioptas2018learning} &  $0.0013$ & $0.052$ \\
3dVAE & $0.0010$ & $0.052$ \\
3dAAE & $0.0009$ & $0.052$ \\
\textbf{3dAAE-G} & $\mathbf{0.0008}$ & $\mathbf{0.051}$ \\
\end{tabular}
\caption{Reconstruction capabilities of the models captured by MMD. Measurements for reconstructions are evaluated on the test split for the considered models trained with EMD loss on training data of the chair class.}
\label{tab:rec}
\end{center}
\end{table}

\begin{figure*}[th!]
\centering
\includegraphics[width=0.9\linewidth]{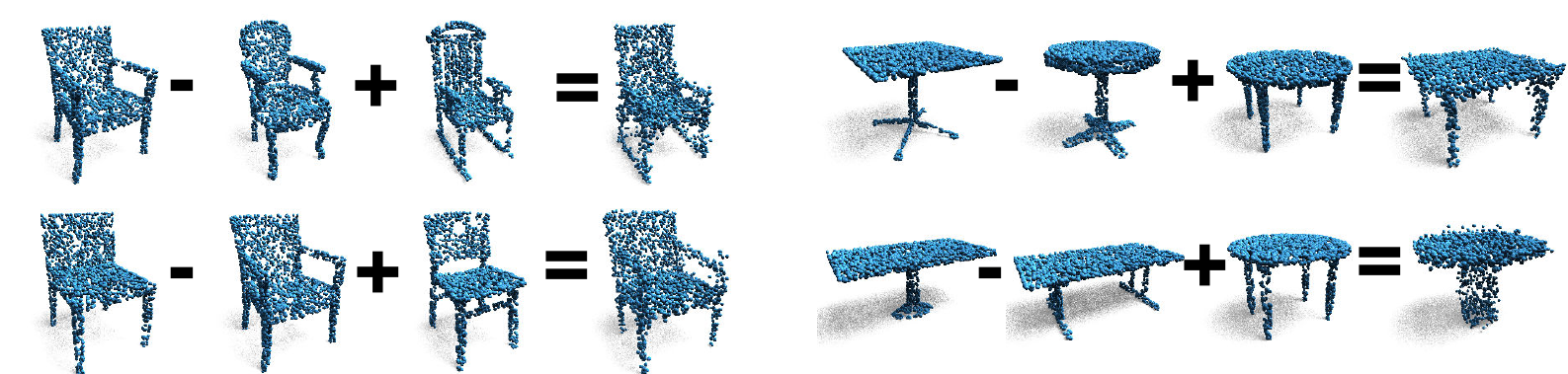}
\caption{Modifying point clouds by performing additive algebra on our latent space encodings by our single-class AAE modes. Top-left sequence: adding rockers to a~chair. Bottom-left: adding armrests to a chair. Top-right: changing table legs from one in the center to four in the corners. Bottom-right: changing table top from rectangle to circle-shaped.}
\label{fig:algebra}
\end{figure*}

\subsection{Generative capabilities}
We present the evaluation of our models, that are involved in sampling procedure directly on latent space $\mathbf{z}$ and compare them with the generative models trained in stacked mode basing on the latent representation of AE. 

For evaluation purposes, we use the chair category and the following five measures: JSD, MMD-CD, MMD-EMD, COV-CD, COV-EMD. For each of the considered models, we select the best one according to the JSD measurement on the validation set. To reduce the sampling bias of these measurements
each generator produces a set of synthetic samples that is 3x the population of the comparative set (test or validation)
and repeat the process 3 times and report the averages.

In Table~\ref{tab:gen_results_1} we report the generative result for memorization baseline model, 5 generative models introduced in \cite{achlioptas2018learning} and three our approaches: 3dVAE, 3dAAE and 3dAAE-G. A baseline model memorizes a random subset of the training data of the same size as the other generated sets. For 3dAAE-G model, we fix the number of Gaussian equal $32$ with different mean values and fixed diagonal covariance matrices.  

It can be observed, that all of our approaches achieved the best results in terms of the JSD measure. Practically, it means that our models are capable of learning better global statistics for generated point locations than reference solutions. We also noticed the slight improvement in MMD criteria for both of the considered distances. In terms of coverage criteria, the results are comparable to the results obtained by the best GAN model and GMM approach. 

In Table~\ref{tab:gen_results_2} we present an additional results on four categories: car, rifle, sofa and table. For further evaluation, we use MMD-EMD and COV-EMD metrics. 3dAAE-G model achieved the best results considering MMD-EMD criterion for each of the datasets and the highest value for three of them. Practically, it means that the assumed Gaussian model with diagonal covariance matrix is sufficient to represent the data in latent space and training GMM in stacked mode is unnecessary when adversarial training with fixed prior is performed. In order to present a qualitative result, we provide synthetic samples generated by the model in Figure~\ref{fig:generated}. 

\setlength{\tabcolsep}{5.5pt}
\begin{table}[h]
\centering
\begin{tabular}{l|c|c|c|c|c}
\multirow{2}{*}{\textbf{Method}} & \multirow{2}{*}{\textbf{JSD}} & \multicolumn{2}{c|}{\textbf{Fidelity}} & \multicolumn{2}{c}{\textbf{Coverage}}\\
& & \textbf{CD} & \textbf{EMD} & \textbf{CD} & \textbf{EMD}\\
\hline
\hline
A & $0.017$ & $0.0018$ & $0.0630$ & $79.4$ & $78.6$\\
\hline
B & $0.176$ & $0.0020$ & $0.1230$ & $52.3$ & $19.0$\\
C & $0.048$ & $0.0020$ & $0.0790$ & $59.4$ & $32.2$\\
D & $0.030$ & $0.0023$ & $0.0690$ & $52.3$ & $19.0$\\
E & $0.022$ & $0.0019$ & $0.0660$ & $67.6$ & $66.9$\\
F & $0.020$ & $0.0018$ & $0.0650$ & $68.9$ & $67.4$\\
\hline
3dVAE & $0.018$ & $\mathbf{0.0017}$ & $0.0639$ & $65.9$ & $65.5$\\
3dAAE & $\mathbf{0.014}$ & $\mathbf{0.0017}$ & $\mathbf{0.0622}$ & $67.3$ & $67.0$\\
\textbf{3dAAE-G} & $\mathbf{0.014}$ & $\mathbf{0.0017}$ & $0.0643$ & $\mathbf{69.6}$ & $\mathbf{68.7}$\\
\end{tabular}
\caption{Evaluating generative capabilities on the test split of the chair dataset on epochs/models selected via minimal JSD on the validation-split. We report A) sampling-based memorization baseline and the following reference methods from~\cite{achlioptas2018learning}: B)~\mbox{r-GAN}, C)~\mbox{l-GAN} (AE-CD), D)~\mbox{l-GAN} (AE-EMD), E)~\mbox{l-WGAN} (AE-EMD), F)~GMM (AE-EMD). The last three rows refer to our approaches: 3dVAE, 3dAAE and 3dAAE-G.}
\label{tab:gen_results_1}
\end{table}
\setlength{\tabcolsep}{6pt}


\begin{table}[h]
\centering
\begin{tabular}{l|r|rr|r|rr}
\multicolumn{1}{c|}{\multirow{2}{*}{\textbf{Class}}} & \multicolumn{3}{c|}{\textbf{Fidelity}} & \multicolumn{3}{c}{\textbf{Coverage}} \\
\cline{2-7} 
\multicolumn{1}{c|}{} & \multicolumn{1}{c|}{PA} & \multicolumn{1}{c}{A} & \multicolumn{1}{c|}{AG} & \multicolumn{1}{c|}{PA} & \multicolumn{1}{c}{A} & \multicolumn{1}{c}{AG} \\
\hline
\hline
car & 0.041 & 0.040 & \textbf{0.039} & 65.3 & 66.2 & \textbf{67.6}\\ 
rifle & 0.045 & 0.045 & \textbf{0.043} & 74.8 & 72.4 &\textbf{75.4}\\
sofa & 0.055 & 0.056 & \textbf{0.053} & \textbf{66.6} & 63.5 & 65.7\\ 
table & \textbf{0.061} & 0.062 & \textbf{0.061} & 71.1 & 71.3 & \textbf{73.0}
\end{tabular}
\caption{Fidelity (\textbf{MMD-EMD}) and Coverage (\textbf{COV-EMD})  metrics on the test split of the car, rifle, sofa and table datasets on epochs/models selected via minimal JSD on the validation split. We report the results for our 3dAAE (denoted as A), 3dAAE-G (denoted by AG) models compared to the reference GMM model (denoted as PA) reported in \cite{achlioptas2018learning}.}
\label{tab:gen_results_2}
\end{table}

\subsection{Latent space arithmetic}
One of the most important characteristics of well-trained latent representations is its ability to generate good-looking samples based on embeddings created by performing interpolation or simple linear algebra. It shows, that model is able to learn the distribution of the data without excessive under- or overfitting \cite{tasse2016shape2vec}. 

In Figure~\ref{fig:interpolations} we show that interpolation technique in the latent space produces smooth transitions between two distinct types of same-class objects. It is worth mentioning that our model is able to change multiple characteristics of objects at once, \eg shape and legs of the table. The similar studies have been performed on binary embeddings obtained in adversarial training with Beta distribution \ref{fig:binary_algebra}. 

Figure~\ref{fig:algebra} presents model ability to learn meaningful encodings that allow performing addition and subtraction in latent space in order to modify existing point clouds. In these examples, we were able to focus on the specific feature that we want to add to our initial point cloud while leaving other characteristics unchanged. Moreover, the operations were done by using only one sample for each part of the transformation.

\subsection{Learning binary embedding with AAE models}
\begin{table}
\centering
\begin{tabular}{c|c|c|c}
\multirow{2}{*}{\textbf{Method}} & \textbf{Continuous} & \multicolumn{2}{|c}{\textbf{Binary}} \\ 
  & Accuracy & Accuracy & mAP \\ 
\hline \hline
Ours 3dAAE$_1$ & 84.68 & 79.82 & 42.94\\ 
Ours 3dAAE$_2$ & 84.35 & 79.78 & 44.09\\ 
AE & 84.85 & 78.12 & 41.76\\ 
\hline
\cite{achlioptas2018learning} & 84.50 & - & - \\
\cite{wu20153d} & 83.3 & - & - \\
\cite{sharma2016vconv} & 75.5 & - & - \\
\cite{girdhar2016learning} & 74.4 & - & - \\
\cite{chen2003visual} & 75.5 & - & - \\
\cite{kazhdan2003rotation} & 68.2 & - & - \\
\end{tabular}
 \caption{Results of point cloud retrieval and embedding classification with Linear SVM on \emph{ModelNet40}. 
 The 3dAAE$_1$ has been trained with constant value of $\lambda=2.0$. 
 In 3dAAE$_2$ we use exponential decay to reduce $\lambda$ as the training proceeds.}
 \label{tab:retrieval_numeric_results}
\end{table}
In the presented work, we have employed the adversarial training to learn informative and diverse binary features. The proposed routine uses $Beta(0.01, 0.01)$ distribution to impose binarization of the embeddings $\mathbf z$ in the training phase, and Bernoulli samples to perform point clouds generation.   

In our studies, we have trained two 3dAAE models and compared their performance with AE to assess the impact of Beta regularization and adversarial training on model quality. The key issue in the experiment is the appropriate calibration of $\lambda$ coefficient, which keeps a balance between reconstruction and adversarial losses. Here we present two top-performing 3dAAE models. The first one (3dAAE$_1$) has been trained with a constant $\lambda=2.0$, which is the best value found in the ablation studies. The latter model (3dAAE$_2$) uses an exponential decay to reduce $\lambda$ as the training proceeds.  

We examine the proposed models in retrieval and embedding classification tasks on ModelNet40. For each experimental setting, we report the corresponding quality metric for continuous and binary embeddings. The binarization is performed simply by taking the threshold value equal to $0$ for AE and $0.5$ for 3dAAE-Beta. 

The results presented in Table \ref{tab:retrieval_numeric_results} provide an evidence that the proposed 3dAAE models classify continuous embeddings with accuracy (Linear SVM) on par with state-of-the-art approaches. Thus imposing additional $Beta(0.01, 0.01)$ prior in adversarial training has no negative impact in this setting. The main benefit arising from 3dAAE training is noticeable once the embeddings are binarized, which leads to superior performance in both retrieval and classification tasks.    

\subsection{Clustering 3D point clouds}
\label{sec:CAAE}

\begin{figure}[ht!]
\begin{center}
    \includegraphics[width=0.9\linewidth,trim={0 11cm 0 0},clip]{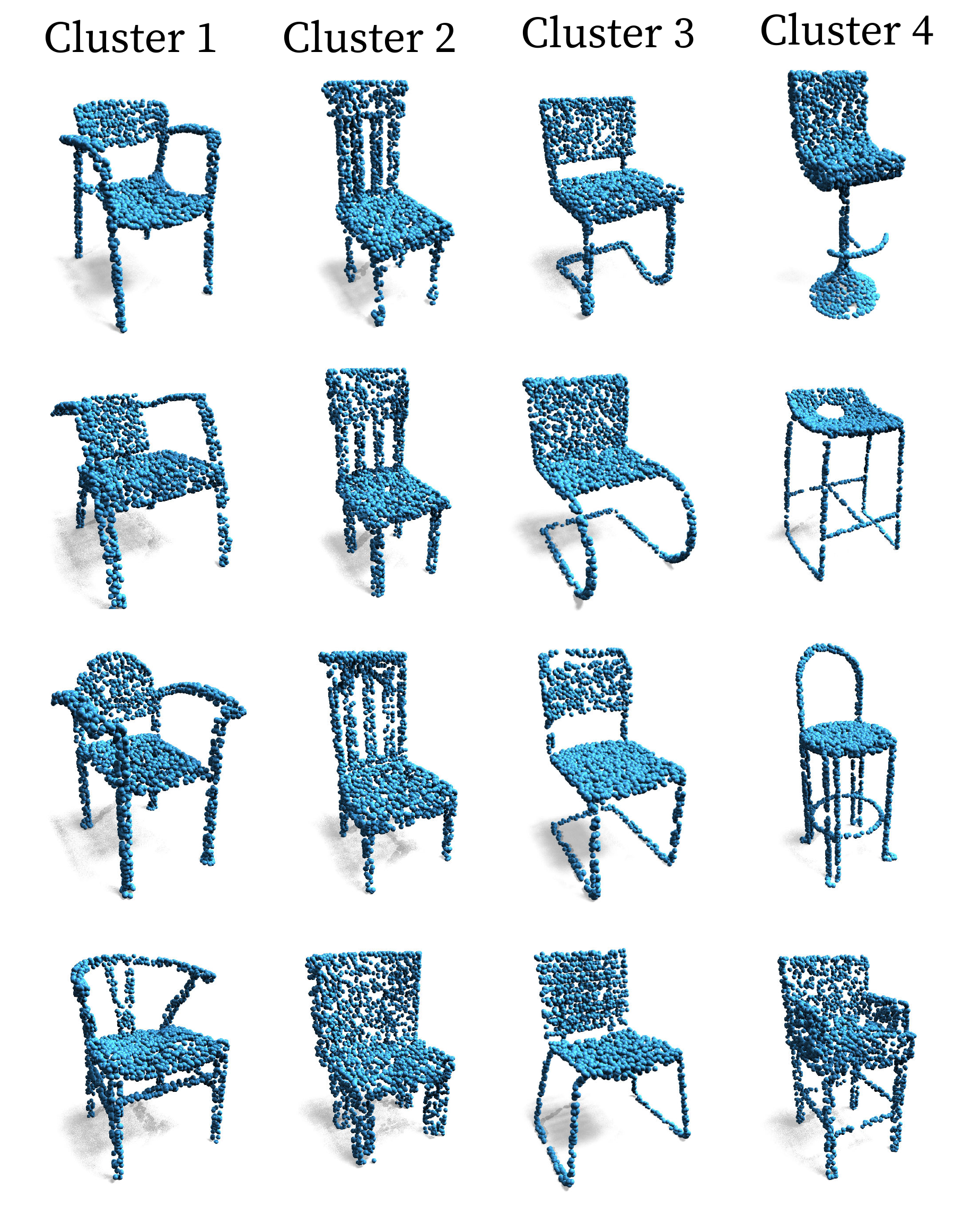}
\end{center}
\caption{Selected examples from test set clustered with adversarial autoencoder with an additional categorical units.}
\label{fig:cluster}
\end{figure}

In this subsection we introduce the extension of the adversarial model for 3D point clouds (\textbf{3dAAE-C}) inspired by \cite{makhzani2015adversarial}, that incorporates an additional one-hot coding unit $\mathbf{y}$. This unit can be interpreted as an indicator of an abstract subclass for the 3D objects delivered on the input of the encoder.   

The architecture for this model extends the existing model (see Figure~\ref{fig:avae}) with an additional discriminator $D_c$. The role of the discriminator is to distinguish between noise samples generated from categorical distribution and one-hot codes $\mathbf{y}$ provided by the encoder. During the training procedure, the encoder is incorporated in an additional task in which it tries to fool the discriminator $D_c$ into creating the samples from the categorical distribution. As a consequence, the encoder is setting $1$ value for the specific subcategories of the objects in the coding space $\mathbf{z}$. 

In Figure~\ref{fig:cluster} we present the qualitative clustering results for the \emph{chair} test data.
We trained the categorical adversarial autoencoder on the chair dataset and set the number of potential clusters to be extracted by our model to $32$. 
We selected $4$ most dominant clusters and presented $6$ randomly selected representatives for each of them. 
It can be observed, that among the detected clusters we can observe the subgroups of chairs with characteristic features and shapes. 

\section{Conclusions}
\label{sec:conclusions}

In this work, we proposed an adversarial autoencoder for generating 3D point clouds. Contrary to the previous approaches \cite{achlioptas2018learning} our model is an end-to-end 3D point cloud generative model that is capable of representing and sampling artificial data from the same latent space. The possibility of using arbitrary priors in an adversarial framework makes model useful not only for representation and generative purposes but also in learning binary encodings and discovering hidden categories. 

To show the many capabilities of our model we provide various of experiments in terms of 3D points reconstruction, generation, retrieval, and clustering. The results of the provided experiments confirm the good quality of the model in the mentioned, tasks that are competitive to the existing state-of-the-art solutions. 

{\small
\bibliographystyle{ieee}
\bibliography{egbib}

\begin{thebibliography}{10}\itemsep=-1pt

\bibitem{achlioptas2018learning}
P.~Achlioptas, O.~Diamanti, I.~Mitliagkas, and L.~Guibas.
\newblock Learning representations and generative models for 3d point clouds.
\newblock In {\em International Conference on Machine Learning}, pages 40--49,
  2018.

\bibitem{ambrogioni2018wasserstein}
L.~Ambrogioni, U.~G{\"u}{\c{c}}l{\"u}, Y.~G{\"u}{\c{c}}l{\"u}t{\"u}rk,
  M.~Hinne, M.~A. van Gerven, and E.~Maris.
\newblock Wasserstein variational inference.
\newblock In {\em Advances in Neural Information Processing Systems}, pages
  2478--2487, 2018.

\bibitem{angelina2018pointnetvlad}
M.~Angelina~Uy and G.~Hee~Lee.
\newblock Pointnetvlad: Deep point cloud based retrieval for large-scale place
  recognition.
\newblock In {\em Proceedings of the IEEE Conference on Computer Vision and
  Pattern Recognition}, pages 4470--4479, 2018.

\bibitem{arjovsky2017wasserstein}
M.~Arjovsky, S.~Chintala, and L.~Bottou.
\newblock {W}asserstein generative adversarial networks.
\newblock In {\em Proceedings of the 34th International Conference on Machine
  Learning}, volume~70 of {\em Proceedings of Machine Learning Research}, pages
  214--223, International Convention Centre, Sydney, Australia, 2017.

\bibitem{bengio2013representation}
Y.~Bengio, A.~Courville, and P.~Vincent.
\newblock Representation learning: A review and new perspectives.
\newblock {\em IEEE transactions on pattern analysis and machine intelligence},
  35(8):1798--1828, 2013.

\bibitem{chang2015shapenet}
A.~X. Chang, T.~Funkhouser, L.~Guibas, P.~Hanrahan, Q.~Huang, Z.~Li,
  S.~Savarese, M.~Savva, S.~Song, H.~Su, et~al.
\newblock Shapenet: An information-rich 3d model repository.
\newblock {\em arXiv preprint arXiv:1512.03012}, 2015.

\bibitem{chen2003visual}
D.-Y. Chen, X.-P. Tian, Y.-T. Shen, and M.~Ouhyoung.
\newblock On visual similarity based 3d model retrieval.
\newblock In {\em Computer graphics forum}, volume 22(3), pages 223--232. Wiley
  Online Library, 2003.

\bibitem{daniels2007robust}
J.~I. Daniels, L.~K. Ha, T.~Ochotta, and C.~T. Silva.
\newblock Robust smooth feature extraction from point clouds.
\newblock In {\em IEEE International Conference on Shape Modeling and
  Applications 2007 (SMI'07)}, pages 123--136. IEEE, 2007.

\bibitem{fan2017point}
H.~Fan, H.~Su, and L.~J. Guibas.
\newblock A point set generation network for 3d object reconstruction from a
  single image.
\newblock In {\em CVPR}, volume 2(4), page~6, 2017.

\bibitem{girdhar2016learning}
R.~Girdhar, D.~F. Fouhey, M.~Rodriguez, and A.~Gupta.
\newblock Learning a predictable and generative vector representation for
  objects.
\newblock In {\em European Conference on Computer Vision}, pages 484--499.
  Springer, 2016.

\bibitem{goodfellow2014generative}
I.~Goodfellow, J.~Pouget-Abadie, M.~Mirza, B.~Xu, D.~Warde-Farley, S.~Ozair,
  A.~Courville, and Y.~Bengio.
\newblock Generative adversarial nets.
\newblock In {\em Advances in neural information processing systems}, pages
  2672--2680, 2014.

\bibitem{gulrajani2017improved}
I.~Gulrajani, F.~Ahmed, M.~Arjovsky, V.~Dumoulin, and A.~C. Courville.
\newblock Improved training of wasserstein gans.
\newblock In {\em Advances in Neural Information Processing Systems}, pages
  5767--5777, 2017.

\bibitem{gumhold2001feature}
S.~Gumhold, X.~Wang, and R.~S. MacLeod.
\newblock Feature extraction from point clouds.
\newblock {\em Proceedings of the 10th international meshing roundtable.
  Newport Beach, California, 2001}, pages 293--305, 2001.

\bibitem{ji20133d}
S.~Ji, W.~Xu, M.~Yang, and K.~Yu.
\newblock 3d convolutional neural networks for human action recognition.
\newblock {\em IEEE transactions on pattern analysis and machine intelligence},
  35(1):221--231, 2013.

\bibitem{kazhdan2003rotation}
M.~Kazhdan, T.~Funkhouser, and S.~Rusinkiewicz.
\newblock Rotation invariant spherical harmonic representation of 3 d shape
  descriptors.
\newblock In {\em Symposium on geometry processing}, volume~6, pages 156--164,
  2003.

\bibitem{kingma2014adam}
D.~P. Kingma and J.~Ba.
\newblock Adam: A method for stochastic optimization.
\newblock {\em arXiv preprint arXiv:1412.6980}, 2014.

\bibitem{kingma2013auto}
D.~P. Kingma and M.~Welling.
\newblock Auto-encoding variational bayes.
\newblock {\em arXiv preprint arXiv:1312.6114}, 2013.

\bibitem{kullback1951information}
S.~Kullback and R.~A. Leibler.
\newblock On information and sufficiency.
\newblock {\em The annals of mathematical statistics}, 22(1):79--86, 1951.

\bibitem{ledig2017photo}
C.~Ledig, L.~Theis, F.~Husz{\'a}r, J.~Caballero, A.~Cunningham, A.~Acosta,
  A.~P. Aitken, A.~Tejani, J.~Totz, Z.~Wang, et~al.
\newblock Photo-realistic single image super-resolution using a generative
  adversarial network.
\newblock In {\em CVPR}, volume 2(3), page~4, 2017.

\bibitem{makhzani2015adversarial}
A.~Makhzani, J.~Shlens, N.~Jaitly, and I.~Goodfellow.
\newblock Adversarial autoencoders.
\newblock In {\em International Conference on Learning Representations}, 2016.

\bibitem{maturana2015voxnet}
D.~Maturana and S.~Scherer.
\newblock Voxnet: A 3d convolutional neural network for real-time object
  recognition.
\newblock In {\em 2015 IEEE/RSJ International Conference on Intelligent Robots
  and Systems (IROS)}, pages 922--928. IEEE, 2015.

\bibitem{pauly2003multi}
M.~Pauly, R.~Keiser, and M.~Gross.
\newblock Multi-scale feature extraction on point-sampled surfaces.
\newblock In {\em Computer graphics forum}, volume 22(3), pages 281--289. Wiley
  Online Library, 2003.

\bibitem{prakhya2015b}
S.~M. Prakhya, B.~Liu, and W.~Lin.
\newblock B-shot: A binary feature descriptor for fast and efficient keypoint
  matching on 3d point clouds.
\newblock In {\em 2015 IEEE/RSJ international conference on intelligent robots
  and systems (IROS)}, pages 1929--1934. IEEE, 2015.

\bibitem{qi2017pointnet}
C.~R. Qi, H.~Su, K.~Mo, and L.~J. Guibas.
\newblock Pointnet: Deep learning on point sets for 3d classification and
  segmentation.
\newblock {\em Proc. Computer Vision and Pattern Recognition (CVPR), IEEE},
  1(2):4, 2017.

\bibitem{qi2017pointnet++}
C.~R. Qi, L.~Yi, H.~Su, and L.~J. Guibas.
\newblock Pointnet++: Deep hierarchical feature learning on point sets in a
  metric space.
\newblock In {\em Advances in Neural Information Processing Systems}, pages
  5099--5108, 2017.

\bibitem{radford2015unsupervised}
A.~Radford, L.~Metz, and S.~Chintala.
\newblock Unsupervised representation learning with deep convolutional
  generative adversarial networks.
\newblock {\em arXiv preprint arXiv:1511.06434}, 2015.

\bibitem{rezende2014stochastic}
D.~J. Rezende, S.~Mohamed, and D.~Wierstra.
\newblock Stochastic backpropagation and approximate inference in deep
  generative models.
\newblock In {\em International Conference on Machine Learning}, pages
  1278--1286, 2014.

\bibitem{Rubner2000}
Y.~Rubner, C.~Tomasi, and L.~J. Guibas.
\newblock The earth mover's distance as a metric for image retrieval.
\newblock {\em International Journal of Computer Vision}, 40:99--121, 2000.

\bibitem{salti2014shot}
S.~Salti, F.~Tombari, and L.~Di~Stefano.
\newblock Shot: Unique signatures of histograms for surface and texture
  description.
\newblock {\em Computer Vision and Image Understanding}, 125:251--264, 2014.

\bibitem{Schonberger17}
J.~L. Sch{\"{o}}nberger, M.~Pollefeys, A.~Geiger, and T.~Sattler.
\newblock Semantic visual localization.
\newblock {\em CoRR}, abs/1712.05773, 2017.

\bibitem{sharma2016vconv}
A.~Sharma, O.~Grau, and M.~Fritz.
\newblock Vconv-dae: Deep volumetric shape learning without object labels.
\newblock In {\em European Conference on Computer Vision}, pages 236--250.
  Springer, 2016.

\bibitem{shoef2019pointwise}
M.~Shoef, S.~Fogel, and D.~Cohen-Or.
\newblock Pointwise: An unsupervised point-wise feature learning network.
\newblock {\em arXiv preprint arXiv:1901.04544}, 2019.

\bibitem{simon2014separable}
T.~Simon, J.~Valmadre, I.~Matthews, and Y.~Sheikh.
\newblock Separable spatiotemporal priors for convex reconstruction of
  time-varying 3d point clouds.
\newblock In {\em European Conference on Computer Vision}, pages 204--219.
  Springer, 2014.

\bibitem{su2015multi}
H.~Su, S.~Maji, E.~Kalogerakis, and E.~Learned-Miller.
\newblock Multi-view convolutional neural networks for 3d shape recognition.
\newblock In {\em Proceedings of the IEEE international conference on computer
  vision}, pages 945--953, 2015.

\bibitem{suger2015traversability}
B.~Suger, B.~Steder, and W.~Burgard.
\newblock Traversability analysis for mobile robots in outdoor environments: A
  semi-supervised learning approach based on 3d-lidar data.
\newblock In {\em 2015 IEEE International Conference on Robotics and Automation
  (ICRA)}, pages 3941--3946. IEEE, 2015.

\bibitem{tasse2016shape2vec}
F.~P. Tasse and N.~Dodgson.
\newblock Shape2vec: semantic-based descriptors for 3d shapes, sketches and
  images.
\newblock {\em ACM Transactions on Graphics (TOG)}, 35(6):208, 2016.

\bibitem{tolstikhin2017wasserstein}
I.~Tolstikhin, O.~Bousquet, S.~Gelly, and B.~Schoelkopf.
\newblock Wasserstein auto-encoders.
\newblock {\em arXiv preprint arXiv:1711.01558}, 2017.

\bibitem{tomczak2018vae}
J.~Tomczak and M.~Welling.
\newblock Vae with a vampprior.
\newblock In {\em International Conference on Artificial Intelligence and
  Statistics}, pages 1214--1223, 2018.

\bibitem{3dgan}
J.~Wu, C.~Zhang, T.~Xue, W.~T. Freeman, and J.~B. Tenenbaum.
\newblock Learning a probabilistic latent space of object shapes via 3d
  generative-adversarial modeling.
\newblock In {\em Advances in Neural Information Processing Systems}, pages
  82--90, 2016.

\bibitem{wu20153d}
Z.~Wu, S.~Song, A.~Khosla, F.~Yu, L.~Zhang, X.~Tang, and J.~Xiao.
\newblock 3d shapenets: A deep representation for volumetric shapes.
\newblock In {\em Proceedings of the IEEE conference on computer vision and
  pattern recognition}, pages 1912--1920, 2015.

\bibitem{xia2014supervised}
R.~Xia, Y.~Pan, H.~Lai, C.~Liu, and S.~Yan.
\newblock Supervised hashing for image retrieval via image representation
  learning.
\newblock In {\em AAAI}, volume 1(2014), page~2, 2014.

\bibitem{yu2018pu}
L.~Yu, X.~Li, C.-W. Fu, D.~Cohen-Or, and P.-A. Heng.
\newblock Pu-net: Point cloud upsampling network.
\newblock In {\em Proceedings of the IEEE Conference on Computer Vision and
  Pattern Recognition}, pages 2790--2799, 2018.

\bibitem{zhou2018voxelnet}
Y.~Zhou and O.~Tuzel.
\newblock Voxelnet: End-to-end learning for point cloud based 3d object
  detection.
\newblock In {\em Proceedings of the IEEE Conference on Computer Vision and
  Pattern Recognition}, pages 4490--4499, 2018.

\bibitem{zieba2018bingan}
M.~Zieba, P.~Semberecki, T.~El-Gaaly, and T.~Trzcinski.
\newblock Bingan: Learning compact binary descriptors with a regularized gan.
\newblock In {\em Advances in Neural Information Processing Systems}, pages
  3612--3622, 2018.

\bibitem{zieba2017training}
M.~Zieba and L.~Wang.
\newblock Training triplet networks with gan.
\newblock {\em arXiv preprint arXiv:1704.02227}, 2017.

\end{thebibliography}
}

\end{document}